\documentclass[10pt,twocolumn,letterpaper]{article}

\usepackage{cvpr}
\usepackage{times}
\usepackage{epsfig}
\usepackage{graphicx}
\usepackage{amsmath}
\usepackage{amssymb}
\usepackage{multirow}
\usepackage{algorithmicx}
\usepackage[ruled,vlined,linesnumbered]{algorithm2e}
\SetAlFnt{\footnotesize}

\usepackage[breaklinks=true,bookmarks=false]{hyperref}

\cvprfinalcopy % *** Uncomment this line for the final submission

\def\cvprPaperID{****} % *** Enter the CVPR Paper ID here

\begin{document}

%%%%%%%%% TITLE
\title{HGAN: Hybrid Generative Adversarial Network}

\author{Seyed Mehdi Iranmanesh\\
West Virginia University\\
{\tt\small seiranmanesh@mix.wvu.edu}
\and
Nasser M. Nasrabadi\\
West Virginia University\\
{\tt\small nasser.nasrabadi@mail.wvu.edu}
}

\maketitle
%\thispagestyle{empty}

%%%%%%%%% ABSTRACT
\begin{abstract}
 In this paper, we present  a simple approach to train Generative Adversarial Networks (GANs) in order to avoid a \textit {mode collapse} issue. Implicit models such as GANs tend to generate better samples compared to explicit models that are trained on tractable data likelihood. However, GANs overlook the explicit data density characteristics which leads to undesirable quantitative evaluations and mode collapse. To bridge this gap, we propose a hybrid generative adversarial network (HGAN) for which we can enforce data density estimation via an autoregressive model and support both adversarial and likelihood framework in a joint training manner which diversify the estimated density in order to cover different modes. We propose to use an adversarial network to \textit {transfer knowledge} from an autoregressive model (teacher) to the generator (student) of a GAN model. A novel deep architecture within the GAN formulation is developed to adversarially distill the autoregressive model information in addition to simple GAN training approach. We conduct extensive experiments on real-world datasets (i.e., MNIST, CIFAR-10, STL-10) to demonstrate the effectiveness of the proposed HGAN under qualitative and quantitative evaluations. The experimental results show the superiority and competitiveness of our method compared to the baselines. 
\end{abstract}

%%%%%%%%% BODY TEXT

\section{Introduction}
Generative models have extensively grown in recent years. The main goal of a generative model is to approximate the true data distribution which is not known. Generative models are based on finding the model parameters that maximize the likelihood of the training data. This is equivalent to minimizing the Kullback-Leibler (KL) divergence ($ D_{KL}(p_{data}||p_{model})$) between the data distribution $p_{data}$ and model distribution $p_{model}$. Although this objective spans multiple modes of the data, it leads to generating vague and undesirable samples~\cite{1}. There are other approaches that minimize $D_{KL} (p_{model}|| p_{data})$ which are usually referred to as the reverse KL divergence~\cite{2} and this is the main idea behind the generative adversarial networks. Although these models generate sharp images, minimizing the reverse KL divergence causes the model distribution to focus on a single mode of the data and ignore the other modes. This is known as the mode collapse in the generative adversarial models~\cite{4}. This happens because the reverse KL divergence measures the dissimilarity between two distributions for the fake samples, and there is no penalty on the fraction of the model distribution that covers the data distribution~\cite{5}. To address this problem, the authors in~\cite{6} suggested the Wassertein distance which has the weakest convergence among existing GANs. However, they used weight clipping to approximate the Wassertain distance which causes a pathological behavior~\cite{4}.   
\begin{figure*}
	\begin{center}
		\includegraphics[width=0.65\linewidth]{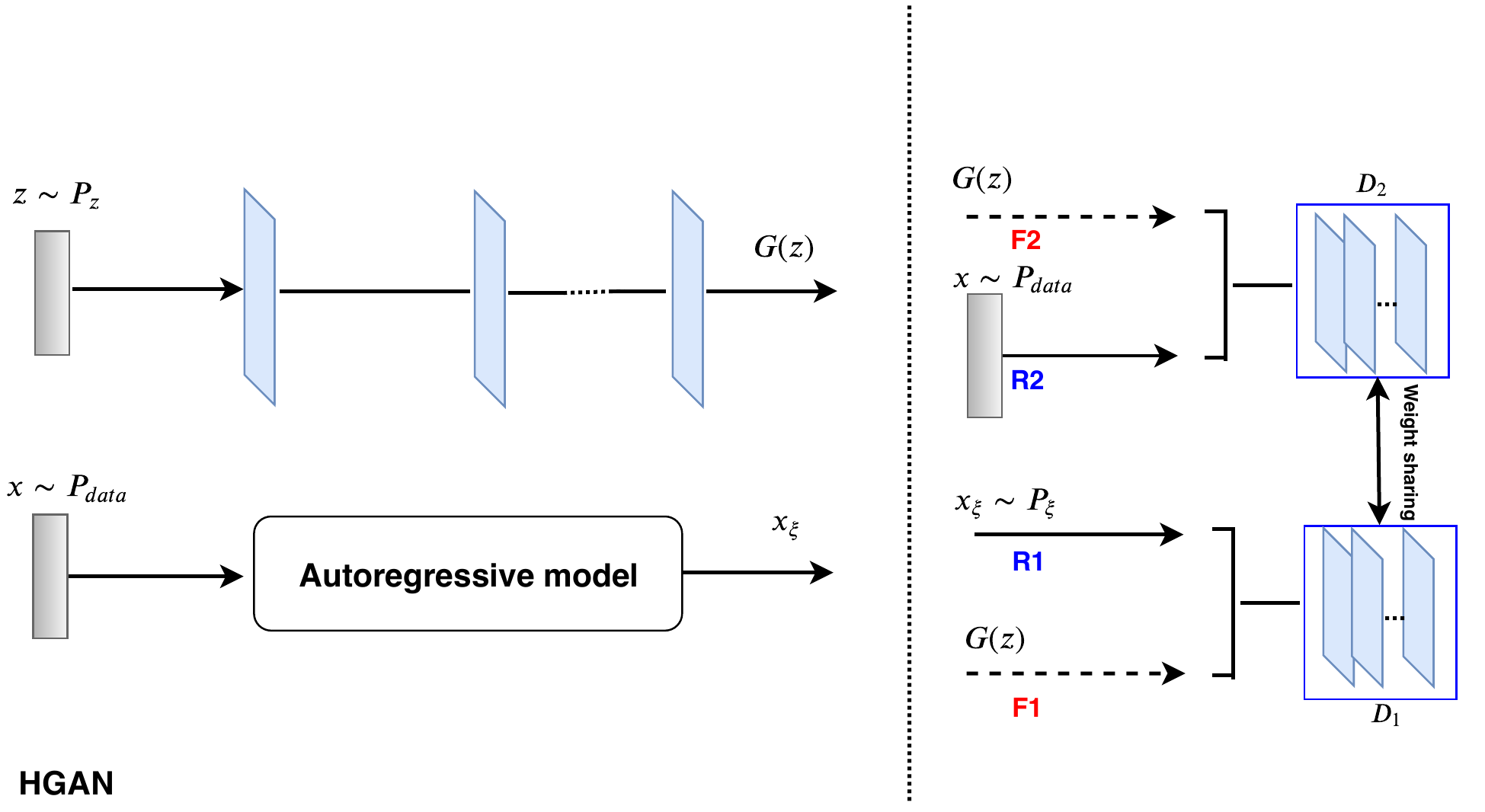}
		
	\end{center}
	\caption{Proposed HGAN framework with an autoregressive model, a generator, and a discriminator is trained by using two types of real data.}
	\label{fig:figure1}
\end{figure*}

In general, the choice for modeling the density function is challenging. There are two ways to estimate the density function namely, implicit methods and explicit methods. Implicit approaches tend to calculate the model parameters without the need for the analytical form of $p_{model}$. Explicit models have the advantage of explicitly calculating the probability densities. There are two well-known implicit approaches, namely GAN and Variational AutoEncoder (VAE) which try to model the data distribution implicitly. The VAEs try to maximize the data likelihood lower bound, while a GAN performs a minimax game between two players during its optimization in which for an optimal discriminator, the algorithm tries to find a generator that minimizes the Jensen-Shannon divergence (JSD). The JSD minimization has been proven empirically to behave more similar to the reverse KL divergence rather than the KL divergence~\cite{7,4}. This behavior leads to the aforementioned problem of mode collapse in GAN models, which causes the generator to create similar looking images with poor diversity of samples. 

In contrast to VAE models which implicitly compute the likelihood of the data space, autoregressive models have the advantage of tractable likelihood and can generate diverse samples. The basic idea of these models is to use the autoregressive connections to model an image pixel by pixel. In fact, autoregressive approaches model the joint distribution of pixels in the image as the product of conditional distributions~\cite{9}. However, these models suffer from a slow synthesis when compared to GANs.
%PixelCNN++~\cite{10} is the most recent autoregressive method that provides a tractable likelihood for the data distribution and generates images with diverse samples.  

The lack of explicit density function in GANs is problematic for two main reasons. Many applications in deep generative models are based on the density estimation. For instance, the count-based exploration methods~\cite{14} rely on density estimation have achieved state-of-the-art performance on reinforcement learning environments~\cite{16}. The second reason is that the quantitative evaluation of the generalization performance of such models is challenging. Since GANs typically are able to generate sharp samples by memorizing the training data, the evaluation criteria based on ad-hoc sample quality metrics~\cite{15} does not capture the mode collapse issue. 

%MGAN~\cite{26} trains many generators by using a classifier and a discriminator. In addition to detecting whether a data sample is fake or real, it predicts which generator produces the sample. This model, minimizes the JSD between the mixture of generators' distribution and data distribution as well as maximizing the JSD among the generators. 

%issue by improving the GAN training process such as in~\cite{miyato2018spectral}. 

Recently some approaches have been trying to solve the mode collapse issue. MGAN~\cite{30} trains many generators by using a classifier and a discriminator. Using many generators and also a classifier in addition to the classical GAN make this model computationally complex and prone to over-fit the training dataset. There are some other approaches that attempt to use autoencoders as regularizers or additional losses to penalize the missing modes~\cite{19,20}. In~\cite{28} authors used an LSTM-based autoregressive model in their discriminator function and considered the reconstruction loss as the penalty for fake data. However, in their GAN model they trained their discriminator only on the true data as it becomes unbounded for the fake data synthesized by the generator.
SNGAN~\cite{miyato2018spectral} utilizes spectral normalization to stabilize the training of discriminator. It controls the Lipschitz constant of the discriminator to mitigate the exploding gradient problem and the mode collapse issue. SAGAN~\cite{zhang2019self} uses the self attention layer to capture the fine details from distant part of image. The authors combined their model with spectral normalization on both the generator and discriminator. BigGAN~\cite{brock2018large} is designed for class-conditional image generation. The focus of the BigGAN model is to increase the number of model parameters and batch size, then configure the model and training process. It utilizes the recent techniques introduced by SNGAN~\cite{miyato2018spectral} and SAGAN~\cite{zhang2019self}. 
StyleGAN~\cite{karras2019style}, is an approach for training generator models capable of synthesizing very large high-quality images via the incremental expansion of both discriminator and generator models from small to large images during the training process. In addition, it changes the architecture of the generator significantly. 

%In~\cite{15}, the authors utilize the mini-batch discrimination trick to allow the discriminator to detect samples that are unusually similar to the other generated samples. This heuristic helps to generate visually more appealing samples at the cost of more computational time. Therefore, this method is usually used in the last hidden layer of the discriminator. Another method is to unroll the optimization of discriminator to make a surrogate objective function in order to help optimizing the generator~\cite{17}. Although their model is robust to mode collapse but it is not clear whether this happens at the cost of losing image quality or not. 

We propose a simple effective GAN architecture and a training strategy with the goal of adversarially distilling the explicit information of the data distribution provided by the autoregressive model in addition to mimicking the real data which leads to generating samples with a distribution very close to the actual data distribution and helps to avoid possible mode collapse. To resolve the issue of sharp good-looking samples but poor likelihood estimation in the case of adversarial learning (and vice versa in the case of maximum likelihood estimation), our proposed hybrid model bridges implicit and explicit learning models by augmenting the adversarial learning with an additional autoregressive model. Our approach combines the implicit and explicit density function estimation into a unified objective function. In our model, the HGAN generator is guided by exploiting the explicit data probability density from the knowledge provided by the autoregressive model while it is also responsible to learn the data distribution via the adversarial learning. HGAN model exploits the complementary statistical properties of data obtained from an autoregressive model by utilizing a GAN to effectively diversify the estimated density function and capturing different modes of the data distribution as well as avoiding possible mode collapse. 
%

%\vspace{-2mm}
In short, our main contributions are: (i) a novel adversarial model to train a generator in a GAN framework in order to stabilize the training process; (ii) the proposed model is able to estimate the data density by mimicking an autoregressive model and simultaneously combining it with the adversarial learning process; (iii) a comprehensive performance evaluation of our proposed method on real-world large-scale datasets of diverse natural scenes as well as mitigating adversarial examples in a defense scenario.   
\vspace{-2mm}
\section{Background}
%In this section, we provide some rudiments of GANs, autoregressive models, and knowledge distillation, necessary to understand the proposed hybrid GAN framework.   
\subsection{Generative Adversarial Nets:}
GAN is a min-max game between a generator G and a discriminator D, both parameterized via neural networks~\cite{3}. Training a GAN can be formulated as the following objective function: 
\begin{align}\label{eq1} 
	&\underset{G}{\operatorname{min}}\, \underset{D}{\operatorname{max}} E_{x\sim P_{data(x)}}[log D(x)]+\\ &E_z\sim P_z [log(1-D(G(z)))] \nonumber ,
\end{align} 
\noindent where $x$ is from a real data distribution $P_{data}$ and $z$ is a sample from a prior distribution $P_z$. The generator is a mapping function from $z$ which approximates $P_{model}$. GAN alternatively optimizes D and G in a minimax game using stochastic gradient-based algorithm. Generator is prone to map every $z$ to a single $x$ that is most probable to be recognized as a true data, and this leads to a mode collapse. Another issue with GAN is that at optimal point of D, minimizing the generator is equal to minimizing the JSD between the true data distribution and model distribution which is empirically shown to cause the mode collapse by generating few modes and ignoring other modes~\cite{7}. 
\subsection{Autoregressive Models:}
Autoregressive models can be designed by using recurrent networks (PixelRNNs) or a CNN (PixelCNNs)~\cite{31}. These models learn the join distribution of pixels of an image $x$ as a product of conditional distributions $p (x_i| x_1, ...,x_{i-1})$, where $x_i$ is a single pixel: 
\begin{align}\label{eq2}
	p(x) =\underset{i=1}{\operatorname{\overset{n^{2}}{\operatorname{\prod}}}} p (x_i| x_1, ...,x_{i-1}) .
\end{align}
The ordering of pixel dependencies is row by row and in each row, pixel by pixel. Therefore, every pixel ($x_i$) depends on all the pixels above and left of it ($x_1, ...,x_{i-1}$). 
\begin{figure*}
	\begin{center}
		\includegraphics[width=0.65\linewidth]{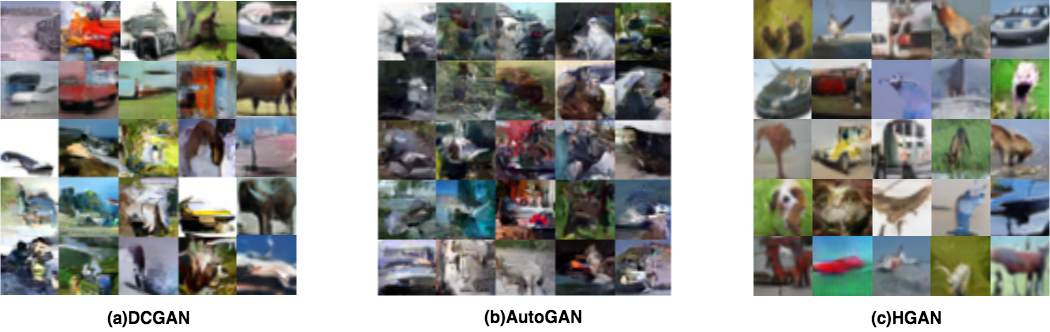}
		
	\end{center}
	\caption{Samples generated by the proposed HGAN compared with the samples generated from DCGAN and AutoGAN on CIFAR-10.} 
	\label{fig:ablation}
\end{figure*}    

\subsection{Knowledge distillation:}

Knowledge distillation is mostly used in image classification problem where the output of neural network is a probability distribution over categories. The probability is calculated by applying a softmax function over \textit {logits} which are the output of the last fully connected layer. Hinton et al.~\cite{47cvpr} used logits to transfer the embedded information in a teacher network to student network. In order to train a student network $F$ to generate student logits $ F(x_i)$, a parameter called temperature $T$ is introduced. Afterwards, the generalized softmax layer converts logits vector $t_i = (t_i^1,..., t_i^C) $ to a probability distribution $q_i$, 

\begin{align}\label{eqdistillation-1} 
M_T(t_i) = q_i,\quad where\quad q_i^j=\dfrac {exp (t_i^j/T)}{\sum_{k} exp(t_i ^k /T)}.   
\end{align}

\noindent where higher temperature $T$ produces softer probability over categories. 

Hinton et al.~\cite{47cvpr} proposed to minimize the KL divergence between teacher and student output as follows: 
\begin{align}\label{eqdistillation-2} 
L_{KD} (F,T) = \dfrac{1}{N}\sum_{i=1}^{N} KL(M_T(t_i)||M_T(F(x_i))).   
\end{align}  

In~\cite {48cvpr} instead of forcing the student to exactly mimic the teacher by minimizing KL-divergence in Equation (\ref{eqdistillation-2}), the knowledge is transferred from teacher to student via discriminator in a GAN-based approach.  
   
\section{Proposed Hybrid GAN:}
We now present our novel hybrid approach to tackle the problem of mode collapse in GANs. In general, GANs can generate good-looking samples but have intractable likelihoods. On the other hand, autoregressive models are likelihood-based generative models which can return explicit probability densities. The idea is to utilize a mixture of these two models rather than a single model as in a typical GAN. 

In our proposed hybrid model, the generator's first task is to learn the data distribution without any explicit model just like a regular GAN such that $G(z) \simeq x \sim P_{data}$. On the other hand, its second task is to perform sampling where it samples a random vector $z\sim P_z$ and maps it to an autoregressive model $P_\xi$ such that $G(z) \simeq x_\xi \sim P_\xi$. This forces our hybrid model to learn the probability density of the autoregressive model using the adversarial training method. These two tasks together provide a hybrid model which gives more attention to the likelihood of the data for estimating $P_{model}$ in the data space.

\begin{algorithm}[t]
	\DontPrintSemicolon
	\SetAlgoLined
	\SetKwInOut{Input}{Input}\SetKwInOut{Output}{Output}
	\Input{minibatch images $x$, number of training batch steps $S$}
	$\theta_\xi, \theta _{G}, \phi_D$ $\gets$ initialize network parameters 
	\BlankLine
	
	\For{$ n = 1$ to ${S}$}{
		$x_\xi \gets p_\xi(x)\,$ \{Forward through auto model \}\;
		$z \sim \mathcal{N}(0,\,1)^Z\,$\{Draw sample of random noise \}\;
		$\hat{x} \gets G(z)\,$ \{Forward through the generator \}\;
		$s_{r_1}\gets D(x_\xi)\,$ \{Auto model output\}\;
		$s_{f_1} \gets D(\hat{x})\,$ \{$G(z)$ output\}\;
		$\hat{x} \gets G(z)\,$ \{Forward through the generator \}\;
		$s_{r_2}\gets D(x_\xi)\,$ \{Real data\}\;
		$s_{f_2} \gets D(x_\xi, \hat{x})\,$ \{$G(z)$ output\}\;
		$\mathcal{L}_D \gets \small{\log} (s_{r_1})+\log (s_{r_2})+ \log (1-s_{f_1}) + \log (1-s_{f_2}) $\;
		$\phi_D \gets \phi_D- \tfrac{\partial\mathcal{L}_D}{\partial \phi_D}$ \{Update discriminator\} \;
		$\mathcal{L}_{G} \gets log (s_{f1})+log (s_{f2}) \,$  \;
		$ \theta_G\gets \theta_G- \tfrac{\partial\mathcal{L}_{G}}{\partial \theta_G}$ \{Update generator\}\;
		
		$\mathcal{L}_{p_\xi} \gets |x_i - p_\xi(x_i)|\,$  \;
		$ \theta_\xi \gets \theta_\xi- \tfrac {\partial\mathcal{L}_{p_\xi}}{\partial \theta_\xi} $ \{Update auto model\}\;
		
	}
\caption{HGAN Training procedure using stochastic gradient descent}
\label{alg1}
\end{algorithm}

A natural question to ask is why one should use adversarial learning when autoregressive model can return a tractable likelihood. The reason is that the synthesis from these autoregressive models are difficult to parallelize and usually inefficient on parallel hardware~\cite{13}. Moreover, it is not practical to perform accurate data manipulation since the hidden layers of autoregressive models have unknown marginal distributions~\cite{16}. However, GAN models are fast in synthesizing and also can have useful latent space for downstream tasks especially in ones which have an encoder such as AGE or ALI~\cite{23,25}. Fig.~\ref{fig:figure1} illustrates the architecture of our proposed Hybrid GAN (HGAN) model.

In naive GAN, the odds that the two distributions $p_g$ and $p_{data}$ share support in high-dimensional space, especially early in training, are very small. If $p_g$ and $p_{data} $ have non-overlapping support the Jensen-Shannon divergence is saturated as is locally constant in $\theta$. Also, there might be a large set of near-optimal discriminators whose logistic regression loss is very close to optimum, but each of these possibly provides very different gradients to the generator. Therefore, training the discriminator might find a different near-optimal solution each time depending on initialization, even for a fixed $g_{\theta}$ and $p_{data}$. We instead employ autoregressive model to augment the gradient information obtained by ordinary back-propagation. In fact, we are interested in manipulating the feature space of a discriminator, using the autoregressive model as a tool to tell us *how* to perform that manipulation.

In our Hybrid GAN, the discriminator observes two types of real inputs: the real data $x$ and the output of the autoregressive model $x_\xi$. The fake input, $G(z)$, is mimicking the output of the autoregressive model $x_\xi \sim P_\xi$ in addition to the real data $x \sim P_{data} $. We consider two terms for the discriminator $D$, namely $D_1$ and $D_2$. $D_1$ is the first discriminator which is related to the first task, where $G(z)$ is fake and $x_\xi$ is real. $D_2$ is related to the second task, where $G(z)$ is fake and $x$ is real. However, all the parameters are shared between discriminators $D_1$ and $D_2$ and in fact there is only one discriminator $D$.
%Therefore, it must implicitly separate the two sources of error: synthetic output of $G(z)$ with the real data $x$, and the synthetic output of $G(z)$ with the output of the autoregressive model $x_\xi$.

%Based on this intuition, which might defuse the learning dynamics, we modified the GAN training algorithm to separate these two mentioned sources of errors. Our approach has two sources of real data, namely $x$ and $x_\xi$ from the autoregressive model $P_\xi$.  Therefore, in addition to the real / fake inputs to the discriminator during the training, we separate the two pairs of fake inputs. These two different pairs of fake input can provide an additional signal to the generator to distinguish between two different sources of error.

\begin{figure}
\begin{center}
	\includegraphics[width=1\linewidth]{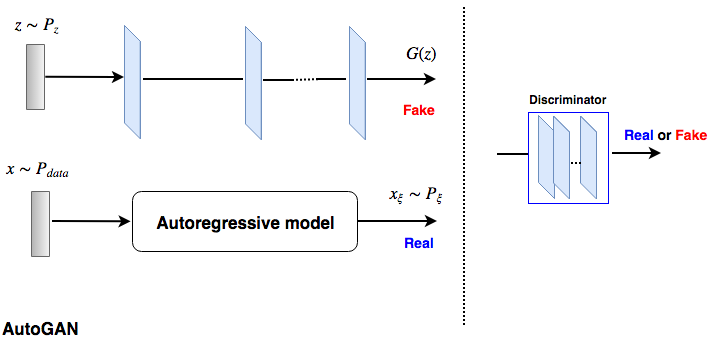}
	
\end{center}
\caption{Training $G(z)$ to mimic the autoregressive model's output with an adversarial learning process. In this network which we denote as AutoGAN, the real data is obtained from the autoregressive model's output and fake data is the generated output from $G(z)$.} 
\label{fig:figure2}
\end{figure}

Algorithm~\ref{alg1} summarizes the training procedure. After getting the input image, output of the autoregressive model, and noise, the proposed model generates a fake image (line 5). $s_{r_1}$ and $s_{r_2}$ indicate the scores for the first ($x_\xi$) and second real inputs ($x$). $s_{f_1}$ and $s_{f_2}$ measures the score of fake inputs containing the generator's output ($G(z)$) trying to mimic the first and second real input, respectively.  Note that we use $\tfrac{\partial\mathcal{L}_D}{\partial \phi_D}$ to indicate the gradient of D's objective function with respect to its parameters, and likewise for $G$ and $p_\xi$. 

In the first fake input of discriminator, the generator attempts to  generate data that is as close as possible to the autoregressive model's output. Therefore, the generator's task is to make $G(z) \simeq x_\xi \sim P_\xi$. However, for the second round of fake input, the generator tries to fool the discriminator in a way that its generated data is as close as possible to the real data. Thus, it is responsible to make $G(z) \simeq  x \sim P_{data} $. While $G$ acts similar to a typical generator in a regular GAN, our hybrid method tries to maximize the likelihood of a mixture model by adversarially distilling the properties of autoregressive model.          

\section {Experiments}  
We show the effectiveness of our proposed approach in different experiments with real-world datasets. For the fair evaluation, we use the same experimental settings that are identical to prior works~\cite{4,38,30}. Therefore, we use the results from the latest state-of-the-art GAN-based models to compare with ours.

\begin{figure*}
	\begin{center}
		\includegraphics[width=0.7\linewidth]{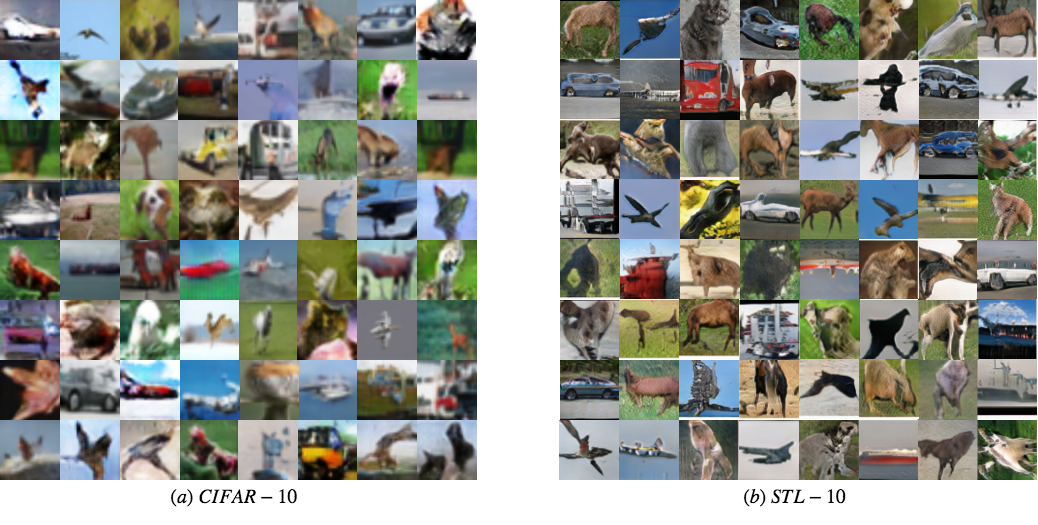}
		
	\end{center}
	\caption{Images generated by our proposed HGAN trained on natural image datasets.} 
	\label{fig:results}
\end{figure*}

We used Pytorch~\cite{33} to implement our framework. The generator and discriminator architecture is adopted from DCGAN~\cite{36}. In addition, pixelCNN++~\cite{10} architecture is chosen for the autoregressive model. For training we used Adam optimizer~\cite{35} with the first-order momentum of 0.5, the learning rate of 0.0002, and batch size of 64. For the generator the ReLU activation~\cite{34}, and for the discriminator the Leaky ReLU activation with the slope of 0.2 is considered. Weights are initialized from an isotropic Gaussian: $\mathcal{N}(0,\,0.01)$ and zero biases.
\begin{table}[]
	\centering
	\caption{Experiment on MNIST dataset containing 10 different modes.}
	\label{table:KLMNIST}
	\scalebox{0.8}{
		\begin{tabular}{|c|c|c|}
			\hline
			GAN Variants    & Chi-square ($\times 10^5 $) & KL Div \\ \hline
			WGAN     & 1.32   & 0.614             \\ \hline
			MIX+WGAN & 1.25   & 0.567             \\ \hline
			DFM      & 1.46   & 0.623             \\ \hline
			Improved-GAN    & 1.13   & 0.436             \\ \hline
			ALI      & 2.34   & 0.875             \\ \hline
			BEGAN    & 1.06   & 0.944             \\ \hline
			MAD-GAN    & 0.24   & 0.145             \\ \hline
			GMAN     & 1.86   & 1.345             \\ \hline
			DCGAN    & 0.90    & 0.322             \\ \hline
			MGAN    & 0.32   & 0.211             \\ \hline
			SAGAN     & 0.29   & 0.148             \\ \hline
			SNGAN     & 0.25   & 0.146             \\ \hline
			\textbf{HGAN}            & \textbf{0.23}   & \textbf{0.141}             \\ \hline
		\end{tabular}}
\end{table}
To show the effectiveness of the proposed framework, we perform two types of experiments on MNIST dataset and compare our methods to the other well-know GANs, namely WGAN~\cite{6}, MIX+WGAN~\cite{15}, DFM~\cite{20}, Improved-GAN~\cite{15}, ALI~\cite{23},  BEGAN~\cite{37}, MAD-GAN~\cite{38}, GMAN~\cite{43}, DCGAN~\cite{36}, MGAN~\cite{30}, SNGAN~\cite{miyato2018spectral}, and SAGAN~\cite{zhang2019self}. It should be noted that our method cannot be compared directly with BigGAN~\cite{brock2018large} and StyleGAN~\cite{karras2019style} since the mentioned models are based on larger models and different settings (i.e., BigGAN is using class conditional setting or StyleGAN purpose is for having more control over the latent space for high resolution image generation). Following~\cite{38}, we reuse the KL-divergence~\cite{39} and the number of captured modes~\cite{40} as the criteria for the comparison to illustrate the superiority of our method compared to others. Moreover, we perform the quantitative experiments on more complicated real-world datasets namely the CIFAR-10~\cite{41} and STL-10~\cite{42} datasets.  
\subsection{MNIST}
The data distribution of the MNIST dataset can be approximated with ten dominant modes. Here, following~\cite{40} we define the term `mode' as a connected component in the data manifold.

\begin{table}[]
\centering
\caption{Results for the Inception scores on CIFAR-10 dataset.}
\vspace{2mm}
\label{table:CIFAR-10}
\scalebox{0.8}{
\begin{tabular}{|c|c|}
	\hline
	Objective  & Inception Score \\ \hline
	DCGAN           & 6.40                   \\ \hline
	AutoGAN          & 6.17                   \\ \hline
	\textbf{HGAN}           & \textbf{7.46}                   \\ \hline
\end{tabular}}
\end{table}

\begin{table}[]
\centering
\caption{Results for the test MODE scores on the MNIST dataset.}
\vspace{2mm}
\label{table:MNIST}
\scalebox{0.8}{
\begin{tabular}{|c|c|}
\hline
Objective  & MODE Score \\ \hline
DCGAN       & 9.28           \\ \hline
AutoGAN     & 9.32           \\ \hline
\textbf{HGAN}        &\textbf{9.51}           \\ \hline
\end{tabular}}
\end{table}

For the sake of evaluation, we train a four-layer CNN classifier on the MNIST digits and then apply it to compute the mode scores in the generated samples from the proposed method. We repeat the procedure and apply the trained classifier to discover the mode scores on different baseline GAN methods. We also have the ground truth by measuring the performance of classifier on the MNIST test set. The number of generated samples from each method is equal to the number of test set which is 10,000. Afterwards, we use Chi-square distance and the KL-divergence to compute distance between the two histograms (ground truth vs. each GAN model). Table~\ref{table:KLMNIST} shows the performance of our proposed HGAN compared to the other methods. From Table~\ref{table:KLMNIST}, it is evident that our proposed method could outperform the other methods in capturing all the modes in the MNIST dataset.

\subsection{Stacked and Compositional MNIST} 
In this experiment, the goal is to explore the performance of our proposed HGAN in a more challenging scenario. In order to illustrate and compare HGAN with other baselines, we utilized similar setup as in~\cite{38}. Authors in~\cite{17} created a Stacked MNIST with 25,600 samples where each sample has three channels stacked together with a random digit from MNIST in each of them. Therefore, the Stacked MNIST contains 1,000 distinct modes in the data distribution. In~\cite{26}, a similar process is applied to MNIST dataset. They created the Compositional MNIST whereby they took three random MNIST digits and placed them at three quadrants of a $64 \times 64$ dimensional image. This also resulted in a data distribution with 1,000 modes. Distribution of the generated samples was estimated with a pre-trained MNIST classifier which classifies the digits in each channel or quadrants, and consequently decides which of the 1,000 modes is generated by the particular GAN method's generator.

\begin{table}[]
\centering
\caption{Stacked-MNIST experiment. There are 1,000 modes in the dataset.}
\label{table:KLMNISTstacked}
\scalebox{0.8}{
\begin{tabular}{|c|c|c|}
\hline
GAN Variants    & KL Div & \# Mode Covered \\ \hline
WGAN     & 1.02   & 868             \\ \hline
MIX+WGAN  & 0.98   & 874             \\ \hline
DFM       & 1.13   & 843             \\ \hline
Improved-GAN    & 1.45   & 847             \\ \hline
ALI       & 2.03   & 802             \\ \hline
BEGAN     & 1.89   & 819             \\ \hline
MAD-GAN     & 0.91   & 890             \\ \hline
GMAN      & 2.17   & 756             \\ \hline
DCGAN     & 2.15    & 712             \\ \hline
MGAN     &  0.94   & \textbf{896}             \\ \hline
SAGAN      & 0.97   & 886             \\ \hline
SNGAN      & 0.91   & 889             \\ \hline
\textbf{HGAN}            & \textbf{0.88}   & 891            \\ \hline
\end{tabular}}
\end{table}

\begin{table}[]
\centering
\caption{Compositional-MNIST experiment. There are 1,000 modes in the dataset.}
\label{table:KLMNISTcompositional}
\scalebox{0.8}{
\begin{tabular}{|c|c|c|}
\hline
GAN Variants    & KL Div & \# Mode Covered \\ \hline
WGAN      & 0.25   & 1000             \\ \hline
MIX+WGAN  & 0.21   & 1000             \\ \hline
DFM       & 0.23   & 965             \\ \hline
Improved-GAN    & 0.67   & 934             \\ \hline
ALI       & 1.23   & 967             \\ \hline
BEGAN     & 0.19   & 999             \\ \hline
MAD-GAN     & 0.074   & 1000             \\ \hline
GMAN      & 0.57   & 929             \\ \hline
DCGAN     & 0.18    & 980             \\ \hline
MGAN     & 0.12   & 1000             \\ \hline
SAGAN      & 0.095   & 1000             \\ \hline
SNGAN      & 0.083   & 1000             \\ \hline
\textbf{HGAN}            & \textbf{0.078}   & \textbf{1000}             \\ \hline
\end{tabular}}
\end{table}

\begin{table}[]
\centering
\caption{Inception scores on the CIFAR-10 and STL-10 datasets. }
\label{table:Inceptionscore}
\scalebox{0.8}{
	\begin{tabular}{|c|c|c|}
		\hline
		Model           & CIFAR-10      & STL-10        \\ \hline
		Real data       & 11.24 $\pm$ 0.16 & 26.08 $\pm$ 0.26 \\ \hline
		WGAN      & 3.82 $\pm$ 0.06  & $-$             \\ \hline
		MIX+WGAN  & 4.04 $\pm$ 0.07  & $-$              \\ \hline
		DFM       & 7.72 $\pm$ 0.13  & 8.51 $\pm$ 0.13  \\ \hline
		Improved-GAN    & 4.36 $\pm$ 0.04  & $-$              \\ \hline
		ALI       & 5.34 $\pm$ 0.05  & $-$              \\ \hline
		BEGAN     & 5.62          &    $-$           \\ \hline
		MAD-GAN     & 7.34          &    $-$           \\ \hline
		GMAN      & 6.00 $\pm$ 0.19  & $-$              \\ \hline
		DCGAN     & 6.40 $\pm$ 0.05  & 7.54          \\ \hline
		MGAN     & \textbf{8.33 $\pm$ 0.10}  & \textbf{9.22 $\pm$ 0.11}          \\ \hline
		SAGAN      & 7.51 $\pm$ 0.15  & 8.61 $\pm$ 0.11  \\ \hline
		SNGAN      & 7.58 $\pm$ 0.12  & 8.79 $\pm$ 0.14  \\ \hline
		\textbf{HGAN}            & 7.46 $\pm$ 0.11  & 8.94 $\pm$ 0.13  \\ \hline
	\end{tabular}}
\end{table}

\begin{table*}[]
	\centering
	
	\caption{FIDs on CIFAR-10 and STL-10 (lower is better).}
	\label{table:FID}
	\scalebox{0.8}{
	\begin{tabular}{|c|c|c|c|c|c|c|c|c|}
		\hline
		Model    & DCGAN & DCGAN+TTUR~\cite{heusel2017gans} & WGAN-GP~\cite{gulrajani2017improved} & GAN-GP & MGAN & SAGAN & SNGAN & HGAN \\ \hline
		CIFAR-10 & 37.7  & 36.9       & 40.2    & 37.7   & 26.7  & 26.3 & \textbf{25.5}  & 26.1  \\ \hline
		STL-10   & -     & -          & 55.1    & -      & -  & 43.6 & 43.2  & \textbf{42.1}  \\ \hline
	\end{tabular}}
\end{table*}

\begin{table*}[]
\centering
\caption{Classification accuracies of using Defense-GAN and Defense-HGAN strategies on the MNIST dataset with $L$ = 200 and $R$ = 10. }
\label{table:AttackMNIST}
\scalebox{0.8}{
	\begin{tabular}{|c|c|c|c|c|}
		\hline
		Attack                  & No Attack (Defense-GAN) & Defense-GAN & No Attack (Defense-HGAN) & Defense-HGAN \\ \hline
		FGSM ($\epsilon = 0.3$) & 0.989                   & 0.961           & 0.991                    & 0.974            \\ \hline
		PGD                     & 0.989                   & 0.956           & 0.991                    & 0.969            \\ \hline
		CW ($l_2$ norm)         & 0.989                   & 0.945           & 0.991                    & 0.965            \\ \hline
	\end{tabular}}
\end{table*}

\begin{figure*}
	\begin{center}
		\includegraphics[width=1\linewidth]{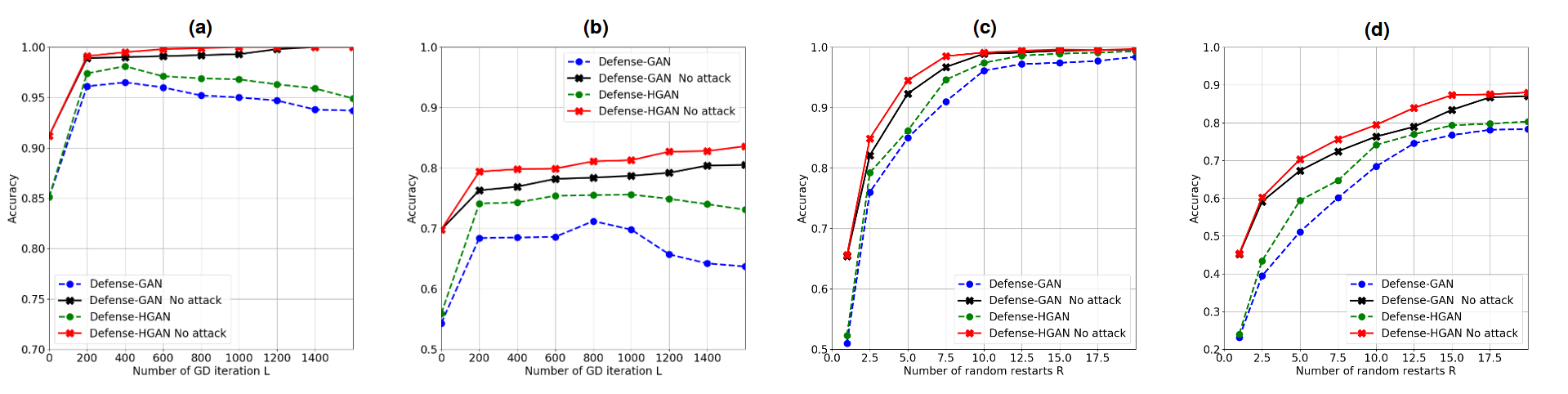}
		
	\end{center}
	\caption{Classification accuracy of Defense-GAN and Defense-HGAN on the MNIST and CIFAR-10 datasets in the case of no attack and also under FGSM white-box attack with $\epsilon=0.3$. (a) MNIST classification accuracy varying $L$ (with $R$ = 10). (b) CIFAR-10 classification accuracy varying $L$ (with $R$ = 10). (c) MNIST classification accuracy varying $R$ (with $L$ = 100). (d) CIFAR-10 classification accuracy varying $R$ (with $L$ = 100).} 
	\label{fig:MNIST_L}
\end{figure*}

\begin{table*}[]
\centering
\caption{Classification accuracies of using Defense-GAN and Defense-HGAN strategies on the CIFAR-10 dataset with $L$ = 200 and $R$ = 10. }
\label{table:AttackCIFAR}
\scalebox{0.8}{
	\begin{tabular}{|c|c|c|c|c|}
		\hline
		Attack                  & No Attack (Defense-GAN) & Defense-GAN & No Attack (Defense-HGAN) & Defense-HGAN \\ \hline
		FGSM ($\epsilon = 0.3$) & 0.763                   & 0.684           & 0.794                    & 0.741            \\ \hline
		PGD                     & 0.763                   & 0.671           & 0.794                    & 0.738            \\ \hline
		CW ($l_2$ norm)         & 0.763                   & 0.646           & 0.794                    & 0.731            \\ \hline
	\end{tabular}}
\end{table*}

Table~\ref{table:KLMNISTstacked} and~\ref{table:KLMNISTcompositional} show the performance of the proposed method as well as other GAN methods in terms of the KL divergence and the number of modes recovered for the Stacked and Compositional MNIST datasets. As shown in Table~\ref{table:KLMNISTstacked}, our method outperformed all the other GAN methods in terms of the KL divergence and the number of captured modes. MGAN surpasses ours in only the number of captured modes. It is evident from Table~\ref{table:KLMNISTcompositional} that our proposed HGAN outperforms all the other baselines in terms of the KL divergence and it is the closest to the true data distribution. Also, in terms of the number of captured modes, our method as well as MGAN, MAD-GAN, WGAN, SNGAN and MIX+WGAN capture all the 1,000 modes in the Compositional MNIST experiment.       

\subsection {Real-world Datasets}

In this section, the proposed HGAN framework is applied on more complicated real-world datasets to evaluate its effectiveness on more challenging large-scale image data. 
\vspace{-2mm}
\subsubsection{Datasets.} We use two widely-adopted datasets, namely CIFAR-10~\cite{41} and STL-10~\cite{44}. CIFAR-10 dataset contains 50,000 training images with the resolution of $32 \times 32$ for 10 different classes: airplane, automobile, bird, cat, deer, dog, frog, horse, ship, and truck. STL-10 dataset subsampled from the ImageNet~\cite{45} and is more diverse database compared to CIFAR-10. This dataset composed of 100,000 images with the resolution of $96 \times 96$. For the sake of fair comparison with the baselines in~\cite{20}, we follow the same procedure as in~\cite{46} to resize the STL-10 down to $48 \times 48$.

\subsubsection {Evaluation Protocols.} For quantitative evaluation, we consider the Inception score which was introduced in~\cite{15}. This metric computes $exp (\mathbb{E}_x[D_{KL}(p(y|x)||p(y))])$, where $p(y|x)$ is the conditional label distribution for image x estimated by the reference Inception model. The metric rewards good and varied samples and is found to be well-correlated to human judgment. The code provided in~\cite{15}, is used to compute the Inception score for 10 partitions of 50,000 generated samples. For qualitative evaluation of the quality of images generated by our proposed HGAN framework, we show the samples generated by HGAN which are drawn randomly rather than cherry-picked. 
\vspace{-2mm}
\subsubsection {Inception Results.} Table~\ref{table:Inceptionscore} shows the Inception scores obtained by our proposed HGAN method as well as the baselines. For the fair comparison, only models which are trained completely in an unsupervised manner without the label information are included in Table~\ref{table:Inceptionscore}. Also, the reported results on STL-10 for DCGAN and D2GAN are based on the models trained on $32 \times 32$ resolution. Table~\ref{table:Inceptionscore} shows the superiority of our proposed HGAN compared to the other methods in the literature for both the STL-10 and CIFAR-10 datasets.

%\vspace{-2mm}
\subsubsection{Image Generation.} For the qualitative assessment, we present samples which are randomly selected from the images generated by the proposed HGAN. It can be seen from Fig.~\ref{fig:results} that the images generated by HGAN are visually recognizable images of cars, ships, trucks, birds, airplanes, dogs, and horses in the CIFAR-10 database. Moreover, in the case of the STL-10 dataset, HGAN is able to produce images including car, trucks, ships, airplanes, and different kinds of animals including horses, cats, monkeys, deers, and dogs with wider range of background such as sky, cloudy sky, sea, and forest. These visually appealing images confirms the diversity of the generated samples by HGAN.

\subsection {Frechet Inception Distance results.} The main disadvantage of the inception score is that it does not compare the statistics of the synthetic samples and the real world ones.  Therefore, we evaluate HGAN using the Frechet Inception Distance (FID) proposed in~\cite{heusel2017gans}. Table~\ref{table:FID} compares the FIDs obtained by HGAN with baselines collected in~\cite{30,miyato2018spectral}. It should be noted that some methods in the literature use the Resnet~\cite{he2016deep} architecture. Here, for the fair comparison we show the results of different methods when using DCGAN architecture.

\subsection{Ablation Study}
In the previous sections, we examined the mode coverage of the proposed framework compared to the other baselines in three separate experiments. In order to show the effectiveness of our HGAN framework, we perform another experiment with two different datasets, namely the MNIST and CIFAR-10 datasets. In this setup, we consider $G(z)$ within two complete separate training approaches. In the first approach, $G(z)$ is trained as a regular GAN such as a DCGAN, and in the second approach $G(z)$ is trained to mimic the autoregressive model's output with an adversarial training. We denote the first approach as DCGAN and the second approach as AutoGAN.  Fig.~\ref{fig:figure2} depicts the framework of AutoGAN. We compare the performance of these two networks with the proposed HGAN in terms of sample quality. 
		
Table~\ref{table:MNIST} and~\ref{table:CIFAR-10} show the highest Inception/MODE scores~\cite{15} of DCGAN, AutoGAN, and HGAN monitored during the training phase. The samples generated by each of the mentioned methods on the CIFAR-10 dataset is also shown in Fig.~\ref{fig:ablation}.

As it is illustrated in Table~\ref{table:MNIST} and~\ref{table:CIFAR-10}, HGAN outperforms both DCGAN and AutoGAN in terms of sample quality. One possible reason behind this is in HGAN, the addition of adversarially distillation of the data information from the autoregressive model (pixelCNN++) in the $G(z)$ objective function can stabilize its optimization, thus avoiding the mode collapse issue. Finally, the hybrid nature of the proposed method leads to a better performance for both datasets. 

\subsection{Comparison with WGAN in Defense Framework}

Despite a very rich research work leading to very interesting GAN algorithms, it is still challenging to assess which algorithm performs better compared to others. In this experiment we evaluate the effectiveness of HGAN compared to WGAN in a defense scenario. We believe this could be another way of assessment for GAN frameworks. 

Adversarial examples~\cite{49cvpr} are neural network inputs which are designed to force misclassification. These inputs often appear normal to humans while cause the neural network to make inaccurate predictions. Various defenses have been proposed to mitigate the effect of adversarial attacks~\cite{50cvpr,54cvpr,55cvpr}. In this experiment we use our proposed HGAN as a defense mechanism against three different white-box attacks: Fast Gradient-Sign Method (FGSM)~\cite{51cvpr}, Carlini-Wagner (CW) attack (with $l_2$ norm)~\cite{52cvpr}, and Projected Gradient Descent (PGD)~\cite{53cvpr}. For the fair comparison, we adopt the same set of experiment as Defense-GAN~\cite{50cvpr}. Instead of using WGAN, we use our proposed HGAN in Defense-GAN framework which we denote as Defense-HGAN. We also compare Defense-HGAN with Defense-GAN in the case of no attack.  Table~\ref{table:AttackMNIST} and \ref{table:AttackCIFAR} show the classification performance of our method compared to Defense-GAN on the MNIST and CIFAR-10 datasets, respectively. It should be noted that the classification accuracy results on the MNIST and CIFAR-10 is 0.994 and 0.886, respectively. We note that Defense-HGAN outperforms Defense-GAN which shows the superiority of our HGAN comparing to WGAN in Defense-GAN framework. 
 
We also compared the effect of different numbers of iterations $L$ and random restarts $R$ for Defense-GAN and Defense-HGAN on the MNIST and CIFAR-10 datasets. Both methods need to look for an appropriate datapoint in the latent space which leads to generating an image closer to the input image. As it is shown in Fig.~\ref{fig:MNIST_L} classification performance of HGAN is better than Defense-GAN which means that HGAN could do a better job in capturing the data distribution compared to WGAN on the MNIST and CIFAR-10 datasets.  
\section{Conclusion}
We have proposed a novel approach to address the mode collapse issue in GANs. Our idea is to design a hybrid model which tries to learn the distribution of data via a mixture of density estimating models utilizing an autoregressive model and an adversarial learning. For this purpose, we introduce a minimax game between a generator, an autoregressive model, and a discriminator to optimize the problem of minimizing the JSD between $P_{data}$ and $P_{model}$. In our proposed HGAN, the generator is responsible to learn the autoregressive model output in addition to modeling the real data just like a regular GAN. Distillation of autoregressive model is beneficial for the HGAN since it also models the distribution of the same data but in an explicit way. It makes the generator to give more attention to the likelihood of the data and stabilize its optimization. This helps the proposed model to capture more data modes which leads to generating a more diversified set of images. 
Comprehensive study on MNIST and also more challenging real-world datasets show the effectiveness of our HGAN in covering data modes and avoiding mode collapse as well as generating diverse and visually appealing images.

{\small
\bibliographystyle{ieee_fullname}
\bibliography{egbib}
}

\end{document}